ARTICLE

# Impact of Missing Values in Machine Learning: A Comprehensive Analysis

Abu-fuad Ahmad[1], Md Shohel Sayeed[2*], Khaznah Alshammari[3] and Istiaque Ahmed[4]

[1,3]New Mexico State University, Las Cruces NM 88001, USA
[2]Multimedia University, Melaka 75450, Malaysia
[4]Osaka Metropolitan University, Osaka 558-8585, Japan

*Corresponding Author: Md Shohel Sayeed. Email: shohel.sayeed@mmu.edu.my

**ABSTRACT**

Machine learning (ML) has become a ubiquitous tool across various domains of data mining and big data analysis. The efficacy of ML models depends heavily on high-quality datasets, which are often complicated by the presence of missing values. Consequently, the performance and generalization of ML models are at risk in the face of such datasets. This paper aims to examine the nuanced impact of missing values on ML workflows, including their types, causes, and consequences. Our analysis focuses on the challenges posed by missing values, including biased inferences, reduced predictive power, and increased computational burdens. The paper further explores strategies for handling missing values, including imputation techniques and removal strategies, and investigates how missing values affect model evaluation metrics and introduces complexities in cross-validation and model selection. The study employs case studies and real-world examples to illustrate the practical implications of addressing missing values. Finally, the discussion extends to future research directions, emphasizing the need for handling missing values ethically and transparently. The primary goal of this paper is to provide insights into the pervasive impact of missing values on ML models and guide practitioners toward effective strategies for achieving robust and reliable model outcomes.

## 1 Introduction

Machine learning, a term widely used across diverse fields, lacks a universally agreed-upon definition due to its broad applicability and the diverse contributions of researchers from various disciplines [1]–[3] This ambiguity is rooted in the extensive areas it covers and the collaborative efforts of researchers with diverse backgrounds. In a broad sense, machine learning can be understood as an algorithmic framework facilitating data analysis, inference, and the establishment of preliminary functional relationships.

The roots of machine learning (ML) can be traced back to the scientific community's interest in the 1950s and 1960s in replicating human learning through computer programs. ML is characterized by its ability to learn from data, improving its performance over time. It extracts knowledge from data for prediction and generating new information, reducing uncertainty by offering guidance on problem-solving, especially in tasks lacking explicit instructions for an analytic solution. ML has proven particularly valuable in image and voice processing, pattern recognition, and complex classification tasks, gaining popularity in both academia and industry due to its superior performance with complex and large-scale data [4]–[9].

ML has become a cornerstone in decision-making across various applications, including healthcare, finance, and environmental monitoring. Its transformative potential lies in discerning patterns, making predictions, and providing insights based on data. The reliability and efficacy of ML models depend on the quality of the datasets used for training and evaluation. However, the pervasive challenge of missing values in datasets, stemming from various sources, necessitates a comprehensive understanding of their impact on ML models. Datasets form the backbone of automated classification or regression systems striving to make informed decisions. Yet, practical datasets from real-world scenarios often exhibit missing values, irregular patterns (outliers), and redundancy across attributes, necessitating solutions for robust model development. Traditionally, missing values are denoted as NaNs (Fig. 1), blanks, undefined, null, or other placeholders [10]–[12] Various factors contribute to missingness, including incorrect data entries, data unavailability, collection issues, missing sequences, incomplete features, file gaps, incomplete information, and more. Irrespective of the causes, handling missing data is essential, as statistical results from datasets with non-random missing values can introduce biases. Furthermore, it is noteworthy that many ML algorithms do not support datasets with missing values [13], [14].

The primary objectives of this paper are to investigate the types and causes of missing values, understand the challenges they pose to machine learning models, explore strategies for handling missing values, and examine the impact on model evaluation metrics. Real-world examples will illustrate the practical implications of addressing missing values in machine learning workflows.

| LotFrontage | LotArea | MasVnrType | MasVnrArea | BsmtQual |
|---|---|---|---|---|
| NaN | 21453 | NaN | 0.0 | TA |
| 67.0 | 5604 | NaN | 0.0 | TA |
| 64.0 | 7301 | BrkFace | 500.0 | NaN |
| NaN | 12692 | NaN | 0.0 | Gd |
| NaN | 2117 | BrkFace | 513.0 | Gd |
| NaN | 8963 | BrkFace | 289.0 | TA |
| NaN | 7000 | BrkFace | 90.0 | TA |
| 35.0 | 4274 | NaN | NaN | Gd |

**Figure 1:** Part of a dataset [42] containing missing values. Missing values are represented as NaN.

The remaining sections of the paper are structured as follows: Section 2 provides a detailed exploration of the different types of missing values and the diverse causes contributing to their occurrence. In Section 3, we present a compilation of an in-depth analysis of the challenges and the various strategies available for effectively handling missing values. Section 4 contains an exploration of how missing values influence common model evaluation metrics and the intricacies introduced in cross-validation and model selection. The subsequent section includes a presentation of case studies and real-world examples illustrating the practical implications of addressing missing values in machine learning workflows. Section 6 contains the results of our analysis, while Section 7 provides several noteworthy observations and recommendations for practitioners dealing with missing values in machine learning. Finally, the last section summarizes the conclusion derived from the study.

## 2 Types and Causes of Missing Values

Missing values can result from various factors such as data collection errors, sensor failures, or participant non-response. Recognizing the causes helps in devising targeted approaches to handle missing values effectively. Fig. 2 shows number of missing values of different features of the diabetes dataset [15]. Understanding the nature of missing values is foundational to devising effective strategies for their handling. There are three key categories of missingness [16], [17] that delineate the mechanism of missingness, namely:

**Missing Completely at Random (MCAR):** In this scenario, the missingness is entirely random and unrelated to any observed or unobserved variables. The missing values are essentially a stochastic result of the data collection process.

**Missing at Random (MAR):** Missingness in this category is related to observed variables but is not dependent on the actual values that are missing. The probability of missing values is conditional on the observed data [18], [19].

**Missing Not at Random (MNAR):** This type of missingness is systematic and related to the unobserved values themselves. In MNAR scenarios, the missing values are directly influenced by the missing data, introducing complexities in handling and imputing missing values.

Understanding these categories is crucial for selecting appropriate imputation techniques and comprehending the potential biases that may arise in the presence of missing values [20], [21].

The causes of missing values in datasets are multifaceted, reflecting the intricacies of real-world data collection processes. Common causes include:

- **Data Collection Errors:** Inaccuracies during data collection processes, such as entry mistakes or misinterpretations, can result in missing values.
- **Sensor Failures:** In applications relying on sensor data, malfunctions or disruptions in sensor operation may lead to missing observations.
- **Participant Non-Response:** In survey or questionnaire-based data collection, participants

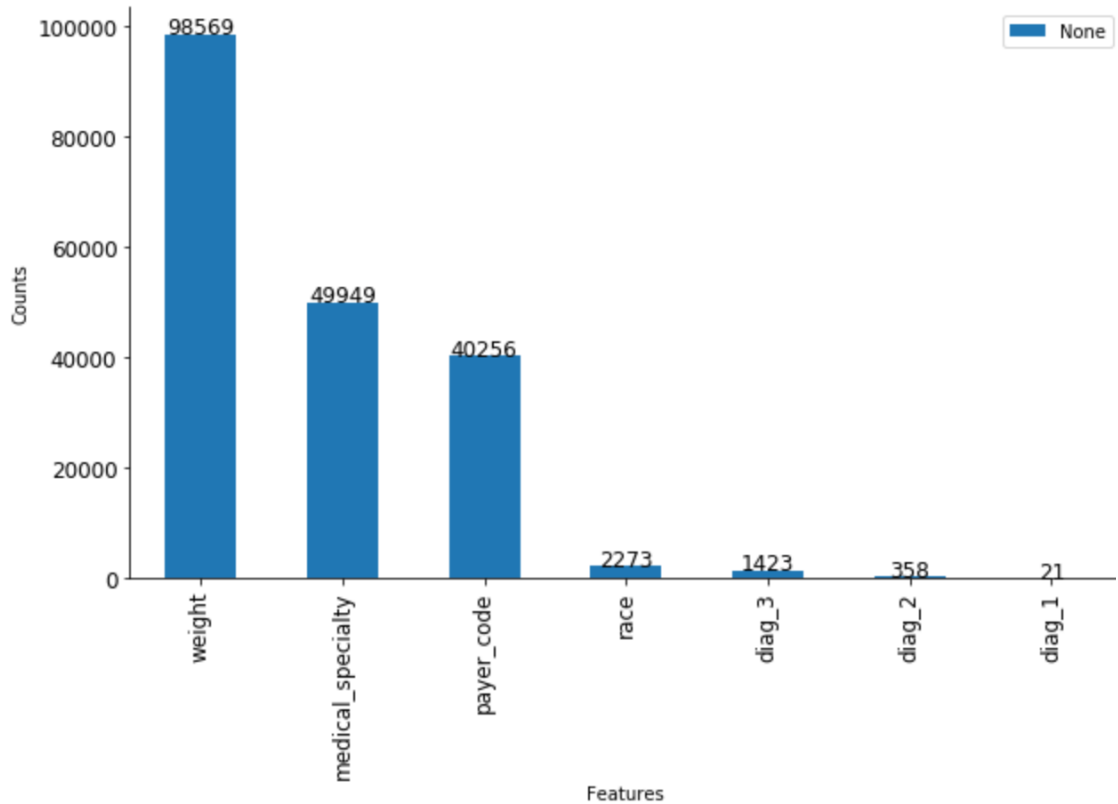

**Figure 1:** Column-wise Missing Values of the Diabetes Dataset [15].

may choose not to respond to certain questions, leading to missing values.
- **Systematic Exclusions:** Certain groups or individuals may be systematically excluded from data collection, leading to missing values that are not random.

Each mechanism presents unique challenges in handling missing data [22], [23]. Therefore, it is essential to identify the mechanism of missingness to select the appropriate strategy to deal with missing values [24], [25]. Acknowledging these causes is essential for formulating targeted strategies for handling missing values and interpreting their impact on machine learning models.

## 3 Strategies for Handling Missing Values

Dealing with missing data is a common challenge in machine learning, especially when the missingness is not completely random. In such cases, the observed data may deviate from a representative sample of the complete dataset, introducing bias that can significantly impact the accuracy and reliability of machine learning models [26]. Addressing this bias becomes crucial to ensure the generalization of models across diverse real-world scenarios. When critical information is missing, the predictive power of machine learning models is significantly reduced, which can impact their ability to identify complex patterns and relationships within the data. This reduction in predictive capability is particularly noteworthy in scenarios where accurate and comprehensive insights are required for effective machine learning applications [24], [25], [27], [28].

Moreover, handling missing values necessitates additional computational resources for imputation or specialized modeling techniques. Whether using statistical methods or machine learning-based approaches, the imputation process demands significant time and resources. This increased computational load can affect the scalability and efficiency of machine learning workflows, particularly in large-scale or real-time applications [29]. To navigate these challenges, one needs to have a thorough understanding of the underlying missing data mechanisms and adopt strategies that aim to mitigate biases, improve predictive capabilities, and manage computational resources effectively.

Handling missing values using imputation methods involves filling in missing values with estimated or predicted values, enabling the inclusion of incomplete observations in the analysis. Several imputation techniques are available, each with its advantages and limitations:

- **Statistics-based Imputation:** Substitute missing values with the mean or median of the observed values for that variable. Sometimes, Missing values are replaced by 0 or a new categorical value. While simple, it assumes that missing values are missing completely at random (MCAR) and may introduce biases [30].
- **Regression-Based Imputation:** Predict missing values based on the relationships observed in the rest of the data. This method is particularly useful when missingness is related to other observed variables [31].
- **K-Nearest Neighbors (KNN) Imputation:** Impute missing values based on the values of their k-nearest neighbors. KNN is effective in capturing local patterns in the data [32] [33].
- **Machine Learning-Based Imputation:** Utilize advanced machine learning algorithms, such as decision trees, random forests, or neural networks, to predict missing values. These methods are flexible and can capture complex relationships [34]–[36].

Alternatively, missing values can be addressed by removing instances or features with missing values [16]. However, the appropriateness of this approach depends on the extent of missingness and its impact on the representativeness of the dataset. Removal may lead to a loss of valuable information and potential biases in subsequent analyses [37].

**4 Impact on Model Performance**

The presence of missing values in a dataset can have a significant impact on the evaluation metrics used to assess the performance of a model. Classification performance after applying imputation techniques on smart building dataset [38] represented in fig. 3. Metrics such as accuracy, precision, recall, and F1-score can be affected, leading to inaccurate assessments of the model's performance [25], [28], [39]. Researchers must be careful when choosing evaluation metrics and adopt strategies to address the bias introduced by missing values [26]. Understanding the impact of missing values on evaluation metrics is crucial for accurate model evaluation.

- **Accuracy:** The ratio of correctly predicted instances to the total instances can be misleading when missing values are present, especially if they are not missing completely at random.

|  | precision | recall | f1-score | support | precision | recall | f1-score | support |
|---|---|---|---|---|---|---|---|---|
| 0 | 0.94 | 1.00 | 0.97 | 1001172 | 1.00 | 1.00 | 1.00 | 37885 |
| 1 | 0.89 | 0.21 | 0.34 | 76999 | 0.95 | 0.94 | 0.94 | 2729 |
| accuracy |  |  | 0.94 | 1078171 |  |  | 0.99 | 40614 |
| macro avg | 0.92 | 0.60 | 0.66 | 1078171 | 0.97 | 0.97 | 0.97 | 40614 |
| weighted avg | 0.94 | 0.94 | 0.92 | 1078171 | 0.99 | 0.99 | 0.99 | 40614 |
| | NaN replaced by 0 | | | | NaN replaced by ML model | | | |

**Figure 2:** Classification Performance after Application of Different Imputation Techniques on Smart Building Dataset [38].

- **Precision:** The proportion of true positive predictions among all positive predictions may be skewed when missing values introduce biases in the data.
- **Recall (Sensitivity):** The ability of the model to capture all relevant instances may be compromised, particularly if certain categories exhibit higher rates of missing values.
- **F1-Score:** The harmonic mean of precision and recall is sensitive to imbalances introduced by missing values, affecting its reliability as a composite metric.

Cross-validation and model selection become more challenging in the presence of missing values. The method chosen for imputing missing data can affect the performance of the model, which means that evaluating models on imputed datasets is essential. The impact of missing values goes beyond just the model selection process. It also affects cross-validation, and it is crucial to handle missing data carefully to ensure unbiased model performance estimation. Furthermore, the method used for imputation can affect model selection, as different methods may lead to variations in model performance [40].

Addressing the influence of missing values on model evaluation metrics is crucial for making informed decisions about the suitability and reliability of machine learning models in the context of specific datasets.

## 5 Case Studies and Real-world Examples

To provide a practical understanding of the impact of missing values, we present case studies and real-world examples across diverse domains. The proposed experiments of missing values was tested with four (4) different datasets: 1) Smart Building System sensor data [38], 2) DataCo supply chain dataset [41] for Big Data Analysis, 3) Diabetics patients readmission prediction of UCI repository [15] and 4) Ames housing price prediction dataset [42]. These examples illustrate how missing values can influence machine learning outcomes and highlight the effectiveness of strategies for handling them.

### *5.1 Healthcare: Predictive Modeling with Clinical Data*

In healthcare, accurate predictions are paramount for patient outcomes. The diabetic dataset is about clinical care records of diabetic patients from 1999 to 2008 in 130 US hospitals and integrated delivery networks. Over 50 feature variables represent patient and hospital outcomes. We explore a case where missing clinical data, such as patient vitals, affected the performance of a predictive model for disease outcomes. Imputation techniques were employed to enhance data completeness, leading to more reliable predictions. Dataset information is presented in the Table 1.

**Table 1:** Diabetic Dataset Properties

| Data Set Characteristics | Multivariate | Number of Instances | 100000 |
|---|---|---|---|
| Attribute Characteristics | Integer | Number of Attributes | 55 |
| Associated Tasks | Classification, Clustering | Missing Values? | Yes |

The diabetic dataset includes patient information such as id number, gender, race, age, etc. and medical histories such as admission type and time, the medical speciality of admitting physician, diagnosis, diabetic medications, lab tests, HbA1c test result, and emergency visits, etc.

## 5.2 Finance: Housing Price Prediction with Incomplete Data

Financial markets rely on accurate predictions for effective decision-making. This housing dataset [42] is about individual residential house sales records between the years of 2006 and 2010 in Ames, Iowa. Eighty (80) feature variables represent the property's quality and quantity. There are 2930 observation records in the dataset. Fig. 4 shows missing values in this dataset by each feature columns.

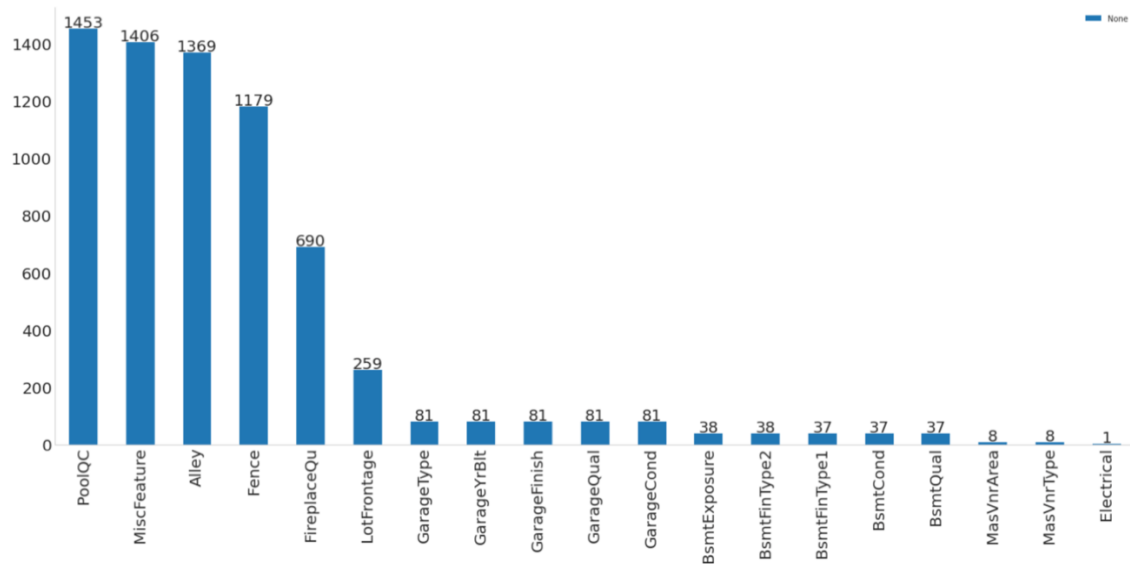

**Figure 4:** Column-wise Missing Values of the Ames Housing Dataset.

Imputation strategies were utilized to address the issue of missing values, which helped to improve the model's ability to capture market trends. The XGBoost model is evaluated by using datasets with imputed missing values using different imputation techniques. When the NAN values were filled with a 0, the model achieved an RMSLE score of 0.14351. On the other hand, using the next valid value on the same column to fill missing values resulted in a score of 0.14348. When the statistical mean of a feature column was used to impute missing values in that column, a performance increment was observed with an RMSLE score of 0.14157.

## 5.3 Environmental Monitoring: IoT Smart Building Big Data

Environmental monitoring involves collecting data on climate variables. We discuss a case where missing values in climate data posed challenges for accurate trend analysis. The experiment took place in the Sutardja Dai Hall (SDH) of UC Berkeley in the year of 2013 [38]. The data was produced by 255 sensors set in 51 different rooms on 4 different floors. Features include different environmental measures such as the humidity of the room, temperature, $CO_2$, and luminosity. The PIR (Passive Infrared Sensor) motion sensors are used to detect the presence of human beings. Data was sampled every 10 seconds for PIR sensors and every 5 seconds for other measurements. This big dataset sized over half (0.50GB) Gigabytes with over 14 Million of instances in CSV format.

Imputation methods were applied to reconstruct missing data points (Fig. 5), enabling more robust assessments of climate patterns. The dataset can be used to find patterns in the smart home environment.

For classification problem on IoT Smart Building dataset, a clean dataset gives an accuracy of 99% compared to 94% on a 0-imputed dataset for missing values.

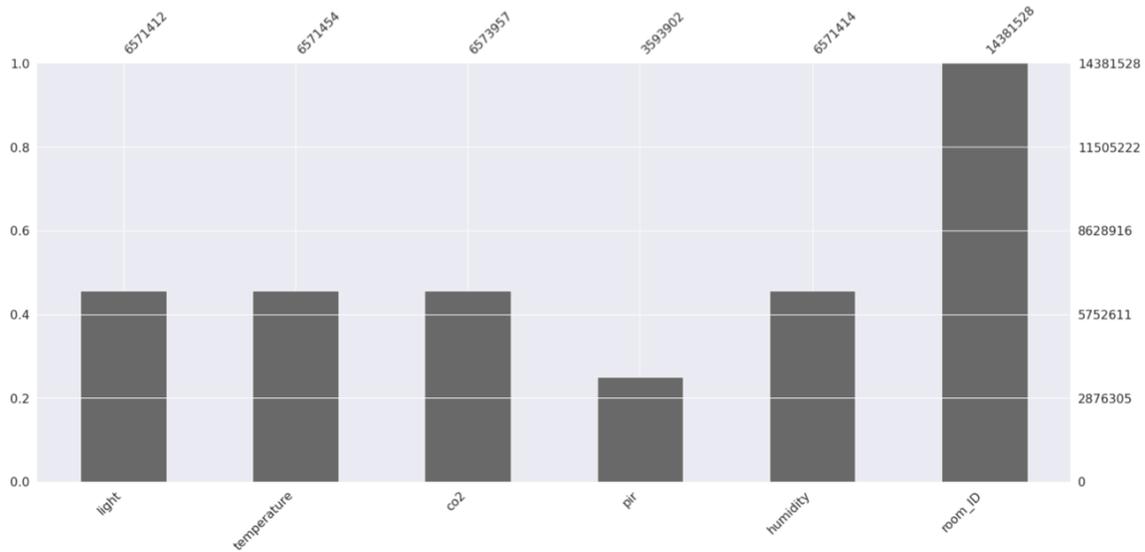

**Figure 5:** Column-wise Missing Values of the Smart Building Dataset.

*5.4 Supply Chain Big Dataset*

The dataset was collected from the DataCo Global company used for the analysis of supply chain management. Machine learning is used in this dataset in the areas of Production, Sales, Provisioning, and Commercial Distribution. This dataset helps in generating knowledge by correlating structured data with unstructured data. At first, there were 180519 data instances and 53 columns. However, after pre-processing and feature engineering, the number of features increased to 3821. The machine learning model trained on this big dataset contains over three (3) thousand features [41]. Analysis results with confusion matrix are shown in fig. 6 and in fig. 7. It is proven that with sufficiently large datasets, ML-based models can outperform the existing methods and also show more stable performance with varying training data sizes (Table 2).

**6 Discussion**

Various experiments were conducted using four different datasets to compare the overall performance of the proposed machine learning (ML) based technique with different models. The ML-based missing value imputation technique outperformed all other traditional imputation methods. The performance of the implemented data augmentation (DA) was compared with the results obtained from different published papers using the same dataset. The mean Root Mean Squared Logarithmic Error (RMSLE) value was calculated based on five trials of the train-test splits while varying the training dataset size from 10% to 90% and the results were plotted in fig. 8. It was observed that the ML-based missing value imputation technique outperformed all other traditional imputation methods. In our experiment, replacing missing values with 0 performed the worst (see Table 2). On the other hand, replacing missing values of any feature column with the median of that column was slightly better than imputing the mean of that feature column.

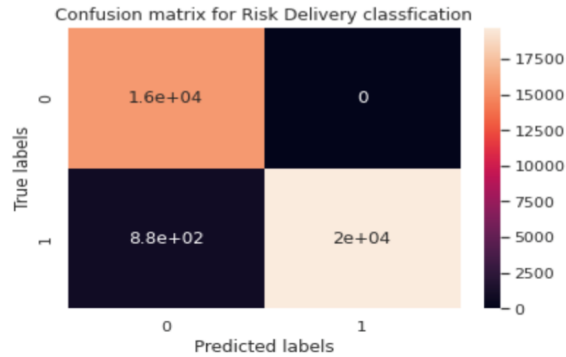

**Figure 3:** Accuracy Results (Confusion Matrix) of Supply Chain Big Data Analytics

**Table 2:** Performance Comparison of ML Model with Existing Methods using MSE Error Metric.

| Training data size | Impute by 0 | Impute by mean | Impute by median | ML Method |
|---|---|---|---|---|
| 10% | 12.0325 | 0.4070 | 0.4067 | 0.1859 |
| 20% | 12.0284 | 0.4155 | 0.4071 | 0.1715 |
| 30% | 12.0258 | 0.417755 | 0.4074 | 0.1641 |
| 40% | 12.0251 | 0.422502 | 0.4135 | 0.1611 |
| 50% | 12.0227 | 0.426761 | 0.4174 | 0.1509 |
| 60% | 12.0282 | 0.431134 | 0.4249 | 0.1498 |
| 70% | 12.0135 | 0.422657 | 0.4091 | 0.1536 |
| 80% | 12.0136 | 0.4319 | 0.4197 | 0.1399 |
| 90% | 12.0296 | 0.3906 | 0.3827 | 0.1145 |

The technique of filling missing data values using machine learning (ML) algorithms is far more effective than traditional methods. ML models such as LinearRegression, DecisionTreeRegressor, LinearSVR, GaussianNB, BaggingRegressor, KNeighborsRegressor, AdaBoostRegressor, and XGBRegressor are used to measure the effectiveness of the missing value imputation technique. The results of the comparison of these models are presented in the Table 3.

It is commonly known that machine learning models become more accurate as the size of the training dataset increases [43]. However, an analysis of Fig. 8 shows that the XGBRegressor and BaggingRegressor models exhibit more consistent and significant improvement patterns. This indicates that, with sufficiently large datasets, the XGBRegressor model outperforms other machine learning methods. Additionally, the XGBRegressor model demonstrates more consistent performance when the size of the training data varies.

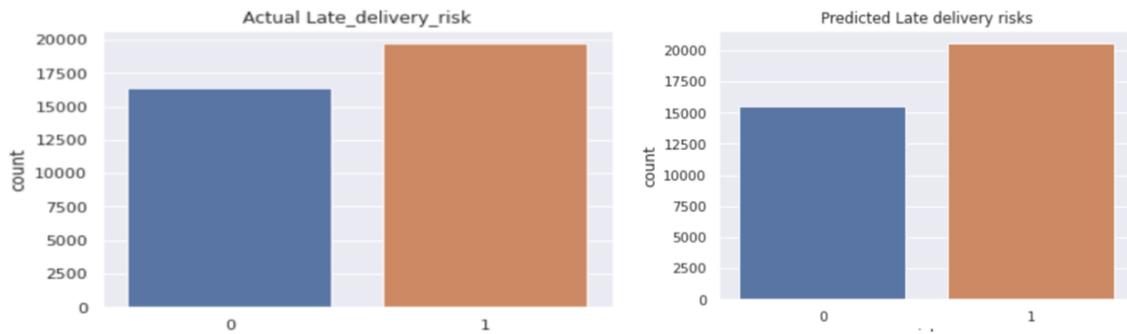

**Figure 4:** Prediction Performance Comparison

These case studies underscore the real-world implications of missing values and demonstrate how strategic handling of missing data can enhance the reliability and effectiveness of machine learning models across diverse domains.

**Table 3:** Performance Comparison of Different Machine Learning (ML) Models using MSE Error

| Train data size % | Linear Reg | Decision Tree | Linear SVR | Gaussian NB | Bagging Reg | KNN | AdaBoost | XGB Regressor |
|---|---|---|---|---|---|---|---|---|
| 10 | 47216.37 | 51765.08 | 48614.98 | 70052.64 | 38461.80 | 59942.25 | 39884.18 | 37222.76 |
| 20 | 46538.26 | 49740.90 | 51191.43 | 65663.78 | 36492.58 | 54714.33 | 39659.44 | 33378.09 |
| 30 | 50161.88 | 43098.24 | 61798.45 | 60245.51 | 34632.06 | 51559.22 | 36054.52 | 31965.93 |
| 40 | 45934.06 | 40668.16 | 64663.69 | 60604.59 | 34601.18 | 49532.64 | 38304.83 | 31786.50 |
| 50 | 37038.73 | 41375.64 | 39158.90 | 57281.71 | 32012.08 | 46358.82 | 37885.17 | 29051.95 |
| 60 | 34306.27 | 39522.03 | 40907.15 | 56233.93 | 30491.38 | 47112.95 | 35614.26 | 27271.41 |
| 70 | 31307.25 | 35678.89 | 60276.58 | 47583.06 | 28013.30 | 43789.97 | 35404.18 | 29030.23 |
| 80 | 31649.75 | 39409.32 | 38281.59 | 48383.08 | 30361.73 | 43543.52 | 33430.26 | 25563.91 |
| 90 | 30139.95 | 35745.35 | 34796.21 | 43092.44 | 24951.43 | 42004.73 | 31667.33 | 22828.05 |

## 7 Observations and Recommendations

In order to select the most appropriate imputation strategy, it is important to carefully consider the missing data mechanism, the characteristics of the dataset, and the desired outcomes of the machine learning task. As the field of machine learning continues to evolve, handling missing data remains a dynamic area of research. Researchers are encouraged to explore innovative approaches that can adapt to the complexities of diverse datasets.

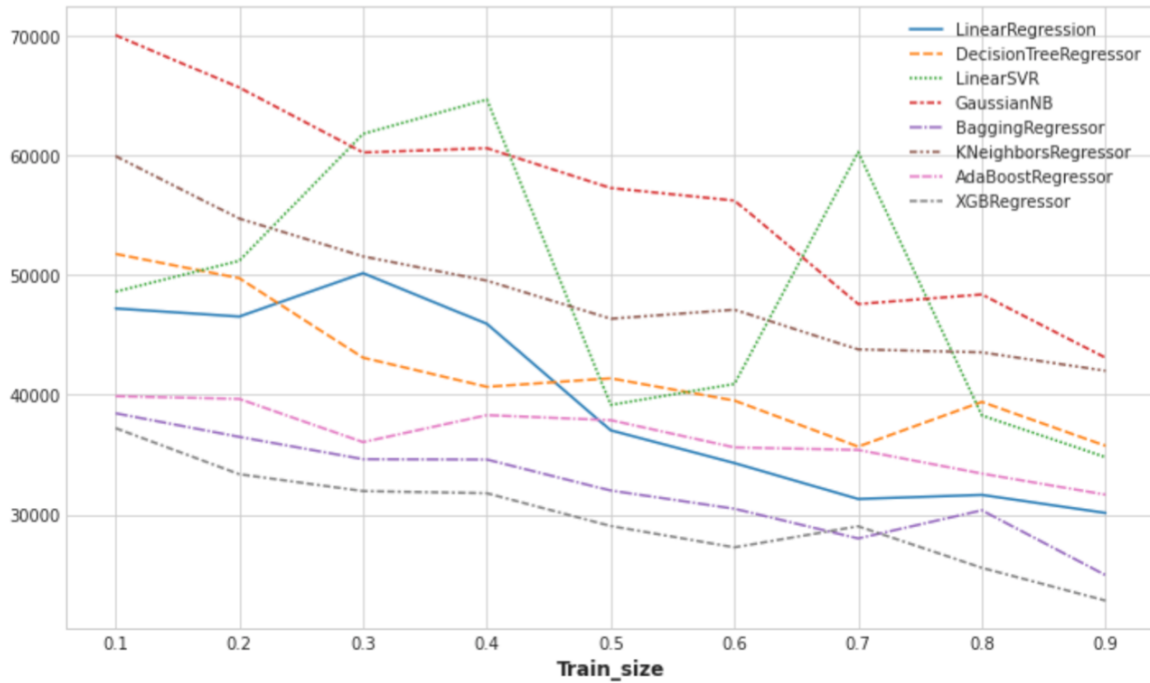

**Figure 5:** Performance trends in ML models on varied training data size measured by MSE.

In this section, we explore future directions and offer recommendations for researchers and practitioners. Our noteworthy observations and recommendations are outlined below:

Continued advancements in imputation techniques, including the integration of deep learning methods, show promise for improving the accuracy and efficiency of missing data handling [44]–[47]. Furthermore, the development of automated tools for selecting and applying appropriate imputation strategies based on dataset characteristics and missing data patterns is an area ripe for exploration. Automated imputation tools can streamline the preprocessing pipeline, making it more accessible to a broader range of practitioners.

Transferring imputation knowledge across domains can help improve the robustness and adaptability of imputation methods. Incorporating domain knowledge into imputation processes can enhance the reliability of imputed values. Future research should explore methods for seamlessly integrating expert knowledge into imputation strategies, especially in fields where contextual understanding is crucial.

To compare imputation methods fairly, benchmark datasets and standardized evaluation protocols are necessary [24], [25], [28]. Most imputation methods assume that the data follows a Gaussian distribution. However, it is necessary to develop techniques that can deal with non-Gaussian data effectively, since there is a wide variety of data distributions in different domains. Researchers are encouraged to contribute to the creation of datasets that capture the complexities of real-world missing data scenarios.

Since missing values can introduce biases [14], [20], [21], ethical considerations when handling and reporting missing data are crucial. Practitioners should prioritize transparency when reporting missing data mechanisms [29]. Additionally, exploring methods that incorporate human expertise in the imputation process can improve the quality of imputed results. Further research should investigate ways to include domain expert feedback and user interactions into the imputation pipeline.

Developing imputation techniques that are aware of biases and ensuring fairness in imputed results is a critical area for future exploration. It is important to address bias in both the imputation process and the downstream effects on machine learning models.

The impact of missing values on common model evaluation metrics requires careful consideration during model development and assessment. Being aware of potential biases and making adjustments to cross-validation and model selection processes are crucial steps in ensuring the reliability and generalization of machine learning models, despite the presence of missing values [26].

Continued research and innovation in these directions will not only advance the field of missing value imputation but also contribute to the overall reliability and effectiveness of machine learning models in the face of incomplete data.

## 8 Conclusion

This article explores the impact of missing values in machine learning and the challenges they pose. To address these challenges, it is important to understand the types and causes of missing values, and to have effective strategies for handling them. Through case studies and real-world examples, the practical implications of addressing missing values in diverse domains are illustrated. It is crucial to handle missing data strategically to enhance the robustness and reliability of machine learning models. The text also outlines future directions and recommendations for researchers and practitioners to further advance the field of missing value imputation. In conclusion, the process of handling missing values in machine learning requires a nuanced understanding of data, a diverse set of imputation tools, and a commitment to ethical and transparent practices. Collaboration between researchers, practitioners, and domain experts will drive innovation and ensure the resilience and effectiveness of machine learning models in the face of incomplete data across a wide range of real-life applications and domains.

**Funding Statement:** The author(s) received no specific funding for this study.

**Author Contributions:** The authors confirm contribution to the paper as follows: study conception and design: A. Ahmad, M. Sayeed, K. Alshammari and I. Ahmed; data collection: A. Ahmad; analysis and interpretation of results: A. Ahmad, M. Sayeed; draft manuscript preparation: A. Ahmad, M. Sayeed, K. Alshammari, I. Ahmed. All authors reviewed the results and approved the final version of the manuscript.

**Availability of Data and Materials:** "The data that support the findings of this study are openly available in [repository name: smartic] at https://github.com/FuadAhmad/smartic . Ames housing dataset used in this research is available at https://www.kaggle.com/c/house-prices-advanced-regression-techniques/data. Diabetes dataset available from UC Irvine Machine Learning repository, https://archive.ics.uci.edu/ml/datasets/Diabetes+130-US+hospitals+for+years+1999-2008#."

**Conflicts of Interest:** The authors declare that they have no conflicts of interest to report regarding the present study.